\documentclass{article}

\PassOptionsToPackage{numbers, compress, sort}{natbib}
\usepackage[preprint]{neurips_2026}

\usepackage[utf8]{inputenc} 
\usepackage[T1]{fontenc}    
\usepackage{hyperref}       
\usepackage{url}            
\usepackage{booktabs}       
\usepackage{amsfonts}       
\usepackage{nicefrac}       
\usepackage{microtype}      
\usepackage[dvipsnames]{xcolor}
\usepackage{graphicx}
\usepackage{pifont}
\usepackage{multirow} 
\usepackage{subcaption}
\usepackage{amsmath,amssymb}

\newcommand{\xmark}{\ding{55}}

\title{MTL-MAD: Multi-Task Learners are Effective Medical Anomaly Detectors}

\author{Bogdan Alexandru Bercean$^{1,3,}$\thanks{Authors contributed equally.}
\And
Florinel Alin Croitoru$^{2,*}$
\And
Vlad Hondru$^{2,*}$
\And
Ciprian Mihai Ceausescu$^{2}$
\And
Andreea Iuliana Ionescu$^{4,5}$
\And
Radu Tudor Ionescu$^{2,}$\thanks{Corresponding author: \texttt{raducu.ionescu@gmail.com}}
\AND
\vspace{-0.5cm}\\
$^1$Rayscape\qquad
$^2$University of Bucharest\qquad
$^3$Polytechnic University of Timi\c{s}oara\\
$^4$Col\c{t}ea Clinical Hospital\qquad
$^5$Carol Davila University of Medicine and Pharmacy
}

\begin{document}

\maketitle

\begin{abstract}
Anomaly detection in medical images is a challenging task, since anomalies are not typically available during training. Recent methods leverage a single pretext task coupled with a large-scale pre-trained model to reach state-of-the-art performance. Instead, we propose to learn multiple self-supervised and pseudo-labeling tasks from scratch, using a joint model based on Mixture-of-Experts (MoE). By carefully integrating multiple proxy tasks, the joint model effectively learns a robust representation of normal anatomical structures, so that anomaly scores can be derived based on how well the multi-task learner (MTL) solves each task during inference. We perform comprehensive experiments on BMAD, a recent benchmark that comprises a broad range of medical image modalities. The empirical results indicate that our multi-task learner is an effective anomaly detector, outperforming all state-of-the-art competitors on BMAD. Moreover, our model produces interpretable anomaly maps, potentially helping physicians in providing more accurate diagnoses. 
\end{abstract}

\setlength{\abovedisplayskip}{2.4pt}
\setlength{\belowdisplayskip}{2.4pt}
\setlength{\abovedisplayshortskip}{2.4pt}
\setlength{\belowdisplayshortskip}{2.4pt}

\section{Introduction}
\vspace{-0.1cm}

Medical anomaly detection from images is a difficult problem, largely due to its inherent open-set nature \cite{Wang-Nature-2023,Bercea-MIDL-2024}. In many clinically important settings, the abnormalities are rare, heterogeneous, and expensive to annotate \cite{Baur-MIA-2021,Kascenas-MIA-2023,Cai-MIA-2025}, whereas large collections of normal or weakly curated data are substantially easier to obtain \cite{Baur-MIA-2021,Bao-CVPRW-2024,Cai-MIA-2025}. This makes unsupervised anomaly detection \cite{Schlegl-IPMI-2017,Schlegl-MIA-2019,Baur-MIA-2021} especially attractive for medical imaging: instead of requiring exhaustive pathology annotations, a model can learn the regularities of normal anatomy and then flag deviations that warrant downstream review. The difficulty, however, is not only the scarcity of positive (anomalous) examples. Medical anomalies vary widely in scale, texture, semantics, and prevalence \cite{Kascenas-MIA-2023,Bao-CVPRW-2024,Cai-MIA-2025}, while the appearance of normal anatomy is itself strongly dependent on organ, modality, acquisition protocol, and dataset construction. Bao et al.~\cite{Bao-CVPRW-2024} make the heterogeneity explicit by introducing BMAD, an unsupervised anomaly detection benchmark spanning six datasets over brain MRI, liver CT, two retinal OCT settings, chest X-ray, and histopathology. The authors found that no single pretext task consistently dominates across settings, and that different approaches exhibit different trade-offs between domains.

\begin{figure}[h!]
\centering
\includegraphics[width=0.98\textwidth]{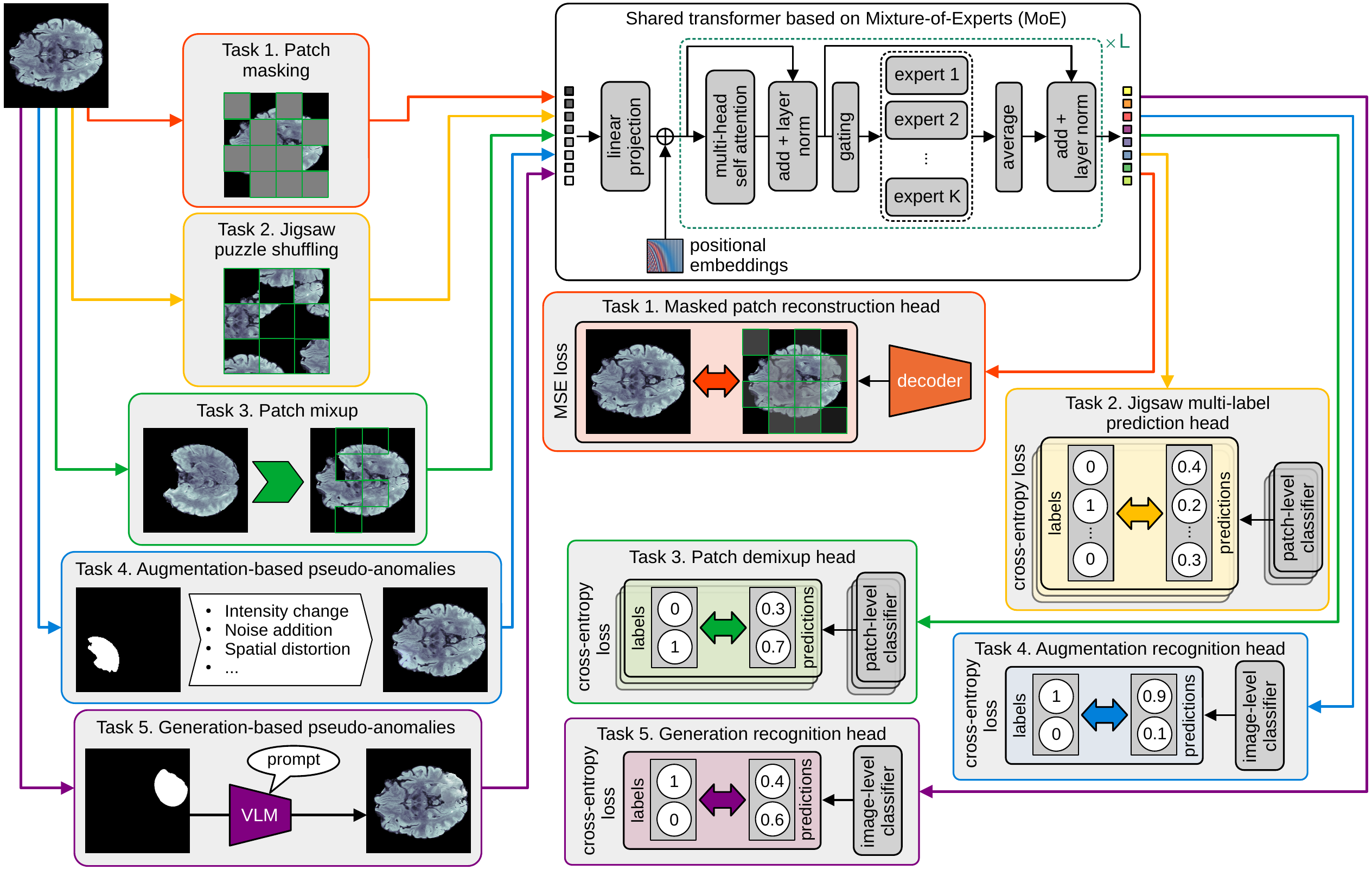}
  \vspace{-0.2cm}
  \caption{Overview of our multi-task learner for medical anomaly detection (MTL-MAD). By meticulously integrating multiple proxy tasks, the joint model learns a robust representation of  normal anatomical structures. During inference, anomaly scores are derived based on how well the multi-task learner solves each task. Positional embeddings are discarded for tokens corresponding to the jigsaw puzzle solving task, to avoid making the task trivial. Best viewed in color.}
  \label{fig:pipeline}
    \vspace{-0.4cm}
\end{figure} 

\vspace{-0.1cm}
Nevertheless, existing unsupervised methods usually rely on a single proxy (pretext) objective \cite{Schlegl-MIA-2019,Li-CVPR-2021,Tan-MLBI-2022,Deng-CVPR-2022,Roth-CVPR-2022}, and, by extension, on a single notion of normality. For instance, reconstruction-based approaches learn to restore or translate images, and then score abnormalities through residuals \cite{Schlegl-IPMI-2017,Schlegl-MIA-2019,Chen-MIDL-2018,Baur-MIA-2021}. Yet, both benchmark analyses and recent medical anomaly detection studies note a recurring limitation: well-trained generators can partially reconstruct pathology and may fail to reflect semantically meaningful or context-level abnormalities in simple pixel-space errors \cite{Heer-MIDL-2021,Bercea-MICCAI-2023,Baur-MIA-2021}. Self-supervised and feature-based alternatives address different weaknesses, yet each proxy objective emphasizes only one failure mode, such as local corruption, distributional deviation, or one-class compactness \cite{Li-CVPR-2021,Tan-MLBI-2022,Roth-CVPR-2022,Deng-CVPR-2022}. The mixed results reported on medical benchmarks are consistent with this picture: no single family of objectives transfers uniformly across modalities, organs, and abnormality types \cite{Bao-CVPRW-2024,Cai-MIA-2025}. 

\vspace{-0.1cm}
To address the limitations of casting medical anomaly detection as a single proxy tasks, we propose MTL-MAD, a multi-task learning (MTL) framework for self-supervised medical anomaly detection (MAD), which combines multi-task learning with task-conditioned Mixture-of-Experts (MoE) \cite{Shazeer-ICLR-2017} inside the transformer backbone. MTL-MAD shares global visual context across tasks, while allowing task-specific specialization through routed feed-forward experts. This design is motivated by prior work on MoE-based multi-task learning \cite{Ma-KDD-2018,Liang-NeurIPS-2022}, showing that task-conditioned routing can disentangle parameter subspaces and mitigate training conflicts that arise under na\"{i}ve shared optimization \cite{Sener-NeurIPS-2018,Ma-KDD-2018,Liang-NeurIPS-2022}. In our setting, different proxy tasks are assigned to different experts or expert subsets, and each task is paired with its own prediction head. As illustrated in Figure \ref{fig:pipeline}, we aggregate a variety of proxy tasks, namely masked image modeling \cite{He-CVPR-2022,Hondru-IJCV-2025,Xie-CVPR-2022}, jigsaw puzzle solving \cite{Noroozi-ECCV-2016,Taleb-IPMI-2021,Wang-ECCV-2022}, synthetic anomaly prediction \cite{Li-CVPR-2021,Tan-MLBI-2022}, and our newly proposed \textsc{DeMixUp}, a patch-level mixing-and-disentanglement objective that mimics local corruptions. These tasks are meticulously integrated to precisely elicit certain abilities during training on normal tissue: masked image modeling induces the ability to generate normal anatomical structures based on neighboring tissue, jigsaw puzzle solving allows the model to understand where certain normal anatomical features are located, \textsc{DeMixUp} makes the model comprehend when a certain structure does not belong to the surrounding context, and pseudo-anomaly detection allows the model to form a rough boundary separating normal samples from everything else. Together, these proxy tasks induce a comprehensive characterization of normal anatomical structures. Moreover, the proposed self-supervised tasks have direct implications towards accurately detecting anomalies, grounded in the following working hypotheses: reconstructing masked patches is more accurate for normal patches than for anomalous patches, rearranging patches with normal tissue is easier than rearranging patch containing abnormal tissue, detecting tissue that differs from surrounding tissue is more obvious when the respective tissue is anomalous, labeling anomalies via a rough boundary around the normality class is easier than having no clue how tight this boundary should be.

\vspace{-0.1cm}
Our key intuition is that the proposed objectives expose complementary abnormality cues that are difficult to capture with any single pretext task. Rather than forcing these signals to collapse into a single score during training, MTL-MAD allows them to be absorbed into the joint backbone, before converting the outputs of the task-specific heads into anomaly scores that are fused at inference by a weighted percentile-rank aggregation scheme. This produces a detector that can leverage heterogeneous evidence without assuming that all tasks should share the same calibration or dominate the final decision in every modality. To validate our hypothesis, we conduct experiments on BMAD~\cite{Bao-CVPRW-2024}, a comprehensive benchmark comprising six medical anomaly detection datasets. Our empirical results indicate that MTL-MAD yields state-of-the-art performance on all six benchmarks, outperforming all competing methods. Unlike state-of-the-art competitors, MTL-MAD achieves high performance without any model pre-training, thereby confirming that its power stems solely from the joint learning of multiple and diverse proxy tasks.

\vspace{-0.1cm}
In summary, our contributions are as follows: 
\begin{itemize}
    \item \vspace{-0.1cm} We propose MTL-MAD, a task-conditioned MoE framework for unsupervised medical anomaly detection within a single ViT backbone.
    \item We introduce \textsc{DeMixUp}, a patch-level self-supervised objective designed to capture local corruption and disentanglement cues complementary to reconstruction and spatial reasoning.
    \item We show that reconstruction, spatial reasoning, local corruption detection, and synthetic anomaly discrimination provide complementary views of normality, when trained jointly, but with specialized experts.
    \item We achieve state-of-the-art results on all six BMAD datasets spanning five medical domains.
\end{itemize}

\section{Related Work}
\vspace{-0.1cm}

Anomaly detection in medical imaging has been explored through several complementary approaches. Reconstruction-based methods~\cite{cai-miccai-2024,sina-jmi-2024,zimmerer-miccai-2019,zhou-tmi-2022,Baur-MIA-2021} train exclusively on normal anatomy, treating deviations in reconstructed outputs as anomaly signals. Recent work has pushed this paradigm forward through diffusion processes~\cite{pinaya-miccai-2022,behrendt-icmidl-2023,baugh-mlim-2025,yuan-mia-2025}. Behrendt et al.~\cite{behrendt-miccai-2024} improved robustness to imperfect reconstructions by sampling multiple pseudo-healthy outputs and measuring pixel-level deviation via the Mahalanobis distance. In contrast, Beizaee et al.~\cite{beizaee-ipmi-2025} employed a masked latent diffusion approach that focuses exclusively on correcting anomalous regions, thereby preserving structural information that is often degraded in full-image diffusion methods. The Q-Former Autoencoder~\cite{dalmonte-wacv-2026} modernized the autoencoder paradigm by coupling frozen encoders based vision foundation models with a learnable bottleneck and perceptual loss, achieving strong unsupervised performance across diverse modalities without any task-specific training of the encoder.

\vspace{-0.1cm}
Vision-language models (VLMs), particularly CLIP, have contributed to the increasing prevalence of few-shot and zero-shot anomaly detectors \cite{zhou-iclr-2024,huang-cvpr-2024,zhang-miccai-2024}. Huang et al.~\cite{huang-cvpr-2024} inserted residual adapters at multiple CLIP encoder stages and combined zero-shot text-based with few-shot memory-bank comparison. Zhang et al.~\cite{zhang-miccai-2024} proposed MediCLIP, a CLIP model fine-tuned on synthetic anomaly generation tasks designed to mimic clinically plausible patterns, achieving substantial gains with less than 1\% of the training set. Shiri et al.~\cite{shiri-miccai-2025} introduced  MadCLIP, a dual-branch adapter architecture with contrastive training under SigLIP loss, explicitly separating normal and abnormal feature subspaces without relying on memory banks or synthetic anomalies. Ma et al.~\cite{ma-cvpr-2025} proposed AA-CLIP to mitigate the intrinsic anomaly-unawareness of CLIP via a two-stage sequential adaptation of both text and visual encoders. 
Beyond binary detection, Zhou et al.~\cite{zhou-miccai-2025} designed UltraAD by extending a few-shot CLIP model to fine-grained multi-class ultrasound classification, incorporating image-aware dynamic prompting and a memory-boosted support module for cross-domain robustness. 

\vspace{-0.1cm}
Unified and cross-domain anomaly detection has emerged as a key frontier for VLMs~\cite{jeong-cvpr-2023,zhou-iclr-2024,gu-cvpr-2025}. Zhu et al.~\cite{zhu-cvpr-2024} developed InCTRL, a single model trained across industrial, medical, and semantic anomaly benchmarks without any target-domain fine-tuning, using few-shot normal images as in-context prompts and computing multi-scale CLIP residuals. The cross-modal generalization setting~\cite{mao-cvpr-2025} requires models trained on known modalities to transfer to entirely unseen ones. Their \emph{transferable visual prototypes} address this by learning modality-agnostic normal and abnormal semantics in visual feature space, unified at inference using visual similarity weighting.

\vspace{-0.1cm}
Multi-task learning offers a complementary mechanism for building generalizable representations. However, dense parameter sharing across tasks frequently induces inter-task interference, where competing gradients corrupt shared weights and degrade overall performance. In natural language processing (NLP), this was addressed through sparsely activated Mixture-of-Experts architectures~\cite{gupta-arxiv-2022} with task-specific routing, directing tokens from different tasks to specialized expert sub-networks, while preserving the computational budget of a dense model. This design provided measurable gains in low-resource transfer and resistance to catastrophic forgetting on multiple NLP tasks. To the best of our knowledge, we are the first to transfer these properties to medical anomaly detection, where a shared model must reconcile heterogeneous supervision signals, across imaging modalities as diverse as MRI, CT, chest X-ray and histopathology. 

\vspace{-0.1cm}
Some self-supervised approaches compose multiple pretext tasks, such as jigsaw, rotation, and re-colorization~\cite{jezequel-tip-2023}, or pair contrastive learning with domain-aware synthetic anomalies~\cite{tian-mia-2023}. However, these methods typically aggregate their signals through independent score fusion at inference or decoupled sequential pipelines~\cite{behzad-miccai-2023}, rather than optimizing all objectives jointly within a single model. Our method adopts a fundamentally different strategy: rather than inheriting representations from pre-trained backbones, adapting foundation models through lightweight modules, or treating tasks as independent components, we propose a joint Mixture-of-Experts model that learns multiple self-supervised and pseudo-labeling objectives simultaneously from scratch, with experts specializing in distinct proxy tasks. Anomaly scores emerge naturally from how poorly each task is solved at inference, additionally leading to interpretable pixel-level anomaly maps that some competing approaches do not provide. 

\vspace{-0.1cm}
Our framework is connected to self-supervised multi-task learning frameworks that have been successful in video anomaly detection \cite{Barbalau-CVIU-2023,Georgescu-CVPR-2021}. While such methods perform joint training via multiple proxy tasks, there are two fundamental differences from our approach. First, the proxy tasks for video anomaly detection are designed to capture normal temporal dynamics and object appearance, while our tasks, e.g.~\textsc{DeMixUp}, are designed to grasp a deep understanding of normal anatomical structures. Second, the shared encoders used in video anomaly detection are based on 2D or 3D convolutions, which may experience difficulties in learning too many, potentially conflicting, proxy tasks, as empirically observed by Barbalau et al.~\cite{Barbalau-CVIU-2023}. In contrast, we train a Mixture-of-Experts framework, in which tasks are routed to different experts, enabling direct task decoupling to avoid conflicting objectives.

\vspace{-0.1cm}
Aside from introducing a multi-task learning framework in medical anomaly detection, we also contribute with a novel proxy task, called \textsc{DeMixUp}. Just as CutPaste \cite{Li-CVPR-2021}, our proxy task is inspired by mixup \cite{Zhang-ICLR-2018} and cutmix \cite{Yun-ICCV-2019} augmentations. Unlike CutPaste \cite{Li-CVPR-2021}, which discriminates between original and altered images\footnote{An altered image contains patches transferred  from another image.}, we classify each patch as pertaining to the original image or the donor one. This enables the model to learn a more contextualized understanding of what constitutes an anomaly within the image, rather than a holistic representation.

\section{Method}
\label{sec_method}
\vspace{-0.1cm}

We propose a multi-task learning framework for unsupervised medical anomaly detection. Our framework is constructed on a Vision Transformer (ViT) backbone \cite{Dosovitskiy-ICLR-2021}, enhanced with Mixture-of-Experts (MoE) layers \cite{Shazeer-ICLR-2017}, where experts specialize themselves on designated proxy tasks. Our model jointly solves three self-supervised pretext tasks: masked image modeling, jigsaw puzzle solving and patch-level \textsc{DeMixUp}, alongside two explicit pseudo-anomaly classification objectives trained on synthetically generated/augmented anomalies. Each task is routed through a dedicated subset of experts, enabling task-specific feature specialization, while retaining a shared attention-based backbone. At inference time, an anomaly score is obtained for each task, and then aggregated into a unified anomaly score. An overview of our approach is illustrated in Figure~\ref{fig:pipeline}.

\subsection{Architecture}
\vspace{-0.1cm}
\noindent
\textbf{Backbone.} We adopt a standard ViT architecture consisting of $L$ transformer encoder layers operating on non-overlapping image patches, i.e.~the input image $X \in \mathbb{R}^{H \times W \times C}$ is split into $N = (H / P)\times(W/P)$ equally-sized patches of $P \times P$ pixels. Each patch is linearly projected to a $d$-dimensional embedding (token). A learnable positional embedding is further added to each token. Each transformer layer consists of a multi-head self-attention (MHSA) block, followed by a feed-forward network (FFN), with layer normalization (LN) and a residual connection:
\begin{equation}
    X^{l} = X^{l-1} + \mathrm{MHSA}(\mathrm{LN}(X^{l-1})), \quad X^l = X^l + \mathrm{FFN}(\mathrm{LN}(X^l)),
\end{equation}
where $X^l$ represents the sequence of tokens (patches) at an arbitrary transformer layer at index $l$.

\vspace{-0.1cm}
\noindent
\textbf{Task-controlled Mixture-of-Experts.} We replace a subset of the standard FFN layers with a MoE module, where the experts are specifically designated to each task. More precisely, unlike standard MoE formulations, our routing is deterministic, such that the tokens corresponding to a task only go through the specifically allocated experts, and the output is computed by taking the average of the assigned experts. Each expert consists of a two-layer feed-forward network with GeLU activations.
During training, we apply expert dropout, randomly dropping experts, but ensuring that at least one expert per task is active. This acts as a regularizer that prevents co-adaption of same-task experts.


\subsection{Proxy Tasks}
\vspace{-0.1cm}
For each proxy task, a custom independent head is attached to the shared encoder. The tasks are carefully chosen and designed such that they empower the model to learn a rich and structured representation of the normal healthy tissue, all of them having complementary capabilities. Combined together, the proposed tasks produce a greater effect: reconstruction-based objectives capture local consistency, permutation-based objectives encode global structure, mixture-based perturbations highlight contextual incompatibilities, and synthetic anomalies provide coarse discrimination. This multi-faceted training paradigm equips the model with both generative and discriminative cues, ultimately enhancing its ability to detect subtle deviations from normal anatomical patterns at inference time.

\vspace{-0.1cm}
\noindent
\textbf{Task 1: Masked image modeling.} Masked image modeling (MIM) \cite{He-CVPR-2022,Hondru-IJCV-2025} consists of inferring masked regions from visible context, thus learning to reconstruct plausible normal anatomical structures. Consequently, abnormal regions will violate the learned reconstruction priors. Starting from the assumption that normal anatomical structures exhibit strong local continuity, when the model is forced to reconstruct the missing regions, it encourages the model to internalize the distribution of healthy tissue. Given $N$ non-overlapping patches $\{x_i\}^N_{i=1}$, a random subset $\mathcal{M} \subset \{1, \dots, N\}$ is masked, and only the visible tokens $\{x_i \mid i \notin \mathcal{M} \}$ are passed through the shared backbone. Using a lightweight transformer decoder with $8$ blocks ($16$ heads with an embedding dimension of $8$), it reconstructs the full set of patches $\hat{x}_i$, with the objective of minimizing the reconstruction error for the masked tokens through the mean squared error (MSE):
\begin{equation}
    \mathcal{L}_{\mathrm{MIM}} (X) = \frac{1}{{\mathcal{|M|}}} \sum_{i \in \mathcal{M}} \left\lVert x_i - \hat{x}_i \right\rVert^2_2.
\end{equation}

\vspace{-0.1cm}
\noindent
\textbf{Task 2: Jigsaw puzzle solving.} While MIM captures local dependencies, it does not explicitly enforce global spatial awareness. To encourage this, we introduce the jigsaw puzzle proxy task, which requires the model to recover the correct spatial arrangement of shuffled patches. Working with many small patches would make the task intractable. Therefore, the first step is to group the $N$ non-overlapping patches $\{x_i\}^N_{i=1}$ into larger tiles, to form a grid of $T \times T$ tiles, such that $T^2 \ll N$, reducing the complexity of the proxy task. A random permutation is applied on the $T^2$ tiles before being fed into the model. The shuffled tiles are encoded via the shared backbone, but not before discarding the positional embeddings, which would make the task trivial. A simple two-layer FFN is trained to predict the original tile positions, for each individual tile. Unlike original formulations for solving jigsaw puzzles \cite{Noroozi-ECCV-2016}, we treat the task as a multi-label classification instead of trying to predict the exact permutation of tiles. For each tile, the model predicts the probability vector $\hat{y}_i \in [0,1]^{T^2}$, not enforcing mutual exclusivity between predictions, e.g.~allowing the model to predict the same position for two different tiles. This formulation avoids over-penalizing the model for small mistakes. We employ the binary cross-entropy loss over tile positions, where $y_{i,j} = 1$ if tile $i$ originates from position $j$, and $0$ otherwise:
\begin{equation}
    \mathcal{L}_{\mathrm{Jigsaw}} (X) = -\frac{1}{{T^2}} \sum^{T^2}_{i}\sum^{T^2}_{j} y_{i,j} \cdot \log (\hat{y}_{i,j}).
\end{equation}

\vspace{-0.1cm}
\noindent
\textbf{Task 3: Patch \textsc{DeMixUp}.} This task sharpens the sensitivity of the model to contextual mismatches, which are often indicative of anomalies. Inspired by mixup \cite{Zhang-ICLR-2018} and cutmix \cite{Yun-ICCV-2019}, patch mixup consists of blending patches from distinct images, then predicting whether each patch token originates from the base or donor image. More precisely, given a base image $X'$ and a donor image $X''$, a composite image $X^{\text{mix}}$ is constructed by replacing a subset of patches $\mathcal{C} \in \{x'_i\}$ with patches at corresponding locations in $\{x''_i\}$. Patches from $\{x^{\text{mix}}_i\}$ are encoded and passed through a classification head based on a two-layer FFN. For each patch $x^{\text{mix}}_i$, the classifier predicts a binary label, indicating whether it originates from the base or donor image. The task is trained with binary cross-entropy loss:
\begin{equation}
    \mathcal{L}_{\textsc{DeMixUp}} (X^{\text{mix}}) = -\frac{1}{N} \sum^N_i y_i \cdot \log \hat{y}_i + (1 - y_i) \cdot \log(1 - \hat{y}_i).
\end{equation}

\vspace{-0.1cm}
\noindent
\textbf{Task 4: Augmentation-based anomaly classification.} We design a pseudo-anomaly classification task, where we employ aggressive image augmentations to produce pseudo-anomalous samples. The aim is to enforce an explicit separation between normal and pseudo-abnormal patterns. When introducing pseudo-anomalies, we provide a weak supervisory signal that helps the model to better approximate a decision boundary around the normal data manifold. One method for generating pseudo-anomalies is applying augmentations on healthy anatomical structures. We first generate anomaly masks by applying unsupervised k-means clustering over pixels, where the number of clusters (superpixels) $k$ is randomly chosen in a given interval. One cluster is randomly chosen for the image manipulation process, which involves operations such as intensity changes, noise addition, spatial distortions, and synthetic artifact generation (opacity and structural defects).
Let $X^{\text{aug}} \in \mathbb{R}^{H \times W \times C}$ denote an augmented image corresponding to an original training image $X$. The classification head is composed of a two-layer FFN that takes the \texttt{[CLS]} token from the backbone as input. The training loss function is the binary cross-entropy:
\begin{equation}
    \mathcal{L}_{\text{Aug-cls}} (Z) = -(y\cdot\log \hat{y} + (1 - y) \cdot\log(1 - \hat{y})), 
\end{equation}
where $Z$ is either an original image $X$, or an augmented image $X^{\text{aug}}$.

\vspace{-0.1cm}
\noindent
\textbf{Task 5: Generation-based anomaly classification.} We introduce an additional pseudo-anomaly classification task, with an analogous classification head and training objective. Different from the augmentation-based pseudo-anomaly generation method, the artificial samples are obtained via generative vision-language models (VLMs), where a healthy sample and a prompt are used to insert an anomaly in the input image, via inpainting. The prompts are specific to the medical domain of the real image that is perturbed. We detail these prompts in the supplementary material. To generate a sample $X^{\text{gen}} \in \mathbb{R}^{H \times W \times C}$, we randomly select a frozen VLM from the following three: Qwen-Image~\cite{Wu-arXiv-2025}, Flux~\cite{Forest-arXiv-2025}, and Stable Diffusion~\cite{Rombach-CVPR-2022}. The classification head is trained via cross-entropy:
\begin{equation}
    \mathcal{L}_{\text{Gen-cls}} (Z) = -(y \cdot\log \hat{y} + (1 - y) \cdot\log(1 - \hat{y})), 
\end{equation}
where $Z$ is either an original image $X$, or a partially generated image $X^{\text{gen}}$. The inpainting masks are generated using overlapping ellipses with randomly assigned quantities, sizes, and center locations. For images dominated by background space, we constrain the ellipse centers to fall within the organ bounds to ensure that the inpainting process targets actual tissue. We emphasize that the random generation and the overlapping of these ellipses naturally produce final masks with varied and complex shapes.


\subsection{Multi-Task Learning} 

\vspace{-0.1cm}
\noindent
\textbf{Curriculum learning / Progressive training.} To improve training stability, we take a different approach from the standard practice: rather than starting with all tasks from the beginning, we integrate the tasks gradually, in a specific order, inspired by curriculum learning \cite{Bengio-ICML-2009,Soviany-IJCV-2022}. During the first phase, only self-supervised tasks (MIM, Jigsaw and \textsc{DeMixUp}) are active, allowing the encoder to learn robust visual representations. In the second phase, the synthetic anomaly detection tasks are activated for joint training, harnessing and further boosting the already learned representations. 

\vspace{-0.1cm}
\noindent
\textbf{Dynamic task weighting.} A common issue in multi-task learning is the conflicting gradients problem \cite{Liu-NeurIPS-2021}. Performance on some tasks might degrade, while attempting to improve performance on others, because their individual task gradients point in opposite directions. To mitigate this problem, we train our model via Fast Adaptive Multitask Optimization (FAMO)~\cite{liu-NeurIPS-2023}, aiming to decrease all losses at an equal relative rate. In our scenario, we define $\mathbf{L}^t= \left[\mathcal{L}_{\mathrm{MIM}}^t, \mathcal{L}_{\mathrm{Jigsaw}}^t, \mathcal{L}_{\textsc{DeMixUp}}^t, \mathcal{L}_{\text{Aug-cls}}^t, \mathcal{L}_{\text{Gen-cls}}^t\right]$ as the vector of loss terms at training iteration $t$. Let $\mathbf{L}_i^t$ denote the loss function at index $i$ from the vector $\mathbf{L}^t$. The final loss function of our joint training pipeline is defined as a weighted sum with terms from this vector, as follows:
\begin{equation}
    \mathcal{L}_t(X, \theta) = c_t \sum_{i=1}^{5} p_i^t \cdot \log{\mathbf{L}^t_i}, 
\end{equation}
where $c_t$ is a normalization constant, $\theta$ are the parameters of the multi-task learner, and $\mathbf{p}^t=\{ p_i^t\}_{i=1}^5$ represent the weights associated to each $\mathbf{L}_i^t$. These weights are computed using the softmax function over a set of learnable logits $\mathbf{p}^t = \mathrm{softmax}(\mathbf{w}^t)$. The update rule of the logits at training iteration $t+1$ is based on the relative loss change from the previous step:
\begin{equation}
    \mathbf{w}^{t+1} =\mathbf{w}^t - \beta \cdot \left(\frac{\partial \mathbf{p}^t}{\partial \mathbf{w}^t}\right)^{\!\!\top}\!\! \cdot \left(\log \frac{\mathbf{L}^t}{\mathbf{L}
    ^{t+1}}\right),
\end{equation}
where $\beta$ is a learning rate and the logits $\mathbf{w^t}$ are all initialized with zero for the first iteration. With this update rule, FAMO forces the weights to automatically shift toward tasks that are not improving fast enough, balancing the relative decrease rates of the individual losses over time.


\subsection{Anomaly Scoring}
\vspace{-0.1cm}

At test time, we compute an anomaly score based on each proxy task, and combine them into a final prediction. For MIM, the score is the mean reconstruction error over the masked patches, while for jigsaw puzzle solving, the cross-entropy serves as the score, in the end both being normalized via $1 - \exp(-x)$. For \textsc{DeMixUp}, we take the sigmoid output as the anomaly probability of each patch, then use the mean of the top-$k$ patches, where $k$ is set to $10$. Similarly, for both pseudo-anomaly classification tasks, the sigmoid output is used directly as an anomaly score.

\section{Experiments}
\label{sec_experiments}
\vspace{-0.1cm}

\noindent
\textbf{Datasets.} We evaluate our method on the BMAD benchmark~\cite{Bao-CVPRW-2024}, which comprises six datasets spanning five medical domains. Specifically, the benchmark includes BraTS2021~\cite{Baid-arXiv-2021} for brain magnetic resonance imaging (MRI) analysis. RESC~\cite{Hu-MIA-2019} and OCT2017~\cite{Kermany-Cell-2018} represent the optical coherence tomography (OCT) domain. The liver computed tomography (CT) domain is represented by a combination of BTCV~\cite{Landman-MICCAI-2015} and LiTs~\cite{Bilic-MIA-2023} datasets. The chest X-ray modality is covered by RSNA~\cite{Wang-CVPR-2017}, and cellular-level analysis is captured through the CAMELYON16~\cite{Ehteshami-JAMA-2017} pathology dataset. We train our pipeline on the official training splits of each of the six datasets, which range in size from 3,000 samples to 26,000 samples. For each training sample, we synthesize two pseudo-anomalies, one via a generative VLM, and the other through classic pixel-level perturbations. We randomly select one of the two pseudo-abnormal images during mini-batch construction.

\vspace{-0.1cm}
\noindent
\textbf{Hyperparameters.} We train the pipeline for $100$ epochs with a batch size of $64$ samples and a starting learning rate of $10^{-3}$. We employ a learning rate scheduler that decays the learning rate down to a minimum of $10^{-6}$ using a factor of $0.5$ when the validation loss does not change significantly. The images are resized to a resolution of $256\times 256$ pixels,
and we use a patch size of $P=16$ in all cases. The hyperparameter configurations of our pipeline can vary between datasets. Specifically, we set the number of transformer encoder blocks $L$ and the number of attention heads within the range of $\left[4, 12\right]$, the hidden dimension $d \in [128, 768]$, and the number of experts $K$ between $5$ and $10$. The exact values are established via grid search on the official validation splits. 

\begin{table}[t]
    \centering
    \caption{Performance comparison of MTL-MAD against leading methods from the literature. The comparison is performed on the entire BMAD benchmark, covering all datasets to demonstrate the consistency of our approach compared to previous state-of-the-art methods. The best score on each dataset is highlighted in {\color{blue}\textbf{blue bold}}, while the second-best score is highlighted in {\color{orange}\underline{orange underline}}. We report average image-level AUROC scores and standard deviations over three runs.}
    \resizebox{0.99\linewidth}{!}{
    \begin{tabular}{l|cccccc}
        \toprule
        Method &  BraTS2021 &  RESC  &   RSNA &  OCT2017 & BTCV+LiTs & CAMELYON16\\ 
        \midrule
         f-AnoGAN~\cite{Schlegl-MIA-2019} & $77.3 \pm 0.18$  & $77.4 \pm 0.85$  &  $55.6 \pm 0.09$ & $73.4 \pm 1.85$ & $58.5 \pm 0.21$ & $69.5 \pm 1.98$ \\
         GANomaly~\cite{Akcay-ACCV-2019}& $74.8 \pm 1.93$ & $52.6 \pm 3.95$ & $62.9\pm 0.65$ & $70.5 \pm 9.98$ & $54.6 \pm 3.06$ & $54.4 \pm 2.57$ \\
         DRAEM~\cite{Zavrtanik-ICCV-2021} & $62.4 \pm 9.03$ & $83.2 \pm 8.21$ & $67.7 \pm 1.72$ & $88.0 \pm 8.36$ & $69.9 \pm 3.86$  & $52.3 \pm 0.77$ \\
        UTRAD~\cite{Chen-NN-2022}  & $82.9 \pm 2.32$  & $89.4 \pm 1.92$ & $75.6 \pm 1.24$ & $96.8 \pm 0.56$ & $55.8 \pm 5.66$ & $69.9 \pm 4.6$ \\ 
         
        DeepSVDD~\cite{Ruff-ICML-2018}  & $87.0 \pm 0.66$ & $74.2 \pm 1.29$  & $64.5 \pm 3.17$ & $76.8 \pm 1.37$ & $53.9 \pm 1.84$ & $60.9 \pm 1.82$\\
        CutPaste~\cite{Li-CVPR-2021}  & $78.8 \pm 0.67$  & $90.2 \pm 0.61$  & $82.6 \pm 1.22$ & $96.8 \pm 0.62$ & $ 59.3 \pm 4.86$ & $75.2 \pm 0.41$\\
        SimpleNet~\cite{Liu-CVPR-2023} & $82.5 \pm 3.34$ & $76.2 \pm7.46$ & $69.1 \pm 1.27$ & $94.7 \pm 2.17$ & {\color{orange}$\underline{72.3 \pm 2.68}$} & $62.4 \pm 3.71$\\
             
        MKD~\cite{Salehi-CVPR-2021} & $81.5 \pm 0.36$  &  $89.0 \pm 0.25$  &  $82.0 \pm 0.12$ & $96.7 \pm 0.26$ & $60.7 \pm 1.19$ & {\color{orange}$\underline{77.5 \pm 0.27}$} \\
             
        RD4AD~\cite{Deng-CVPR-2022} &  $89.5 \pm 0.91$ &$87.8 \pm 0.87$ &  $67.6 \pm 1.11$ & $97.3 \pm 0.79 $ & $60.4 \pm 1.17$ & $66.8 \pm 0.71$ \\
            
        STFPM~\cite{Yamada-arXiv-2022}  & $83.0 \pm 0.67$ &  $84.8 \pm 0.50$  & $72.9 \pm 1.96$ & $96.8 \pm 0.23$ & $61.8 \pm 1.58$ & $66.4 \pm 1.01$\\
        PaDiM ~\cite{Defard-ICPR-2021} & $ 79.0 \pm 0.38 $& $75.9 \pm0.54$  & $77.5 \pm 1.87$ & $91.8 \pm 0.96$ & $50.8 \pm 0.61$ & $67.3 \pm 0.32$  \\
        PatchCore~\cite{Roth-CVPR-2022}  & $91.7 \pm 0.36$  &  $91.6 \pm0.10$  & $76.1 \pm 0.67$ & {\color{orange}$\underline{98.6 \pm 0.03}$} & $60.3 \pm 0.76$ & $69.3 \pm 0.21$\\
        CFA~\cite{Lee-IEEE-2022}  & $84.4 \pm 0.87$ & $69.9 \pm 0.26$ &  $66.8 \pm 0.23$ & $79.5 \pm 0.56$ & $62.0 \pm 1.08$& $65.6 \pm 0.10$\\
        CFLOW~\cite{Gudovskiy-WACV-2022} & $74.8 \pm 5.32$  & $75.0 \pm 5.81$  & $71.5 \pm 1.49$  & $85.4 \pm 2.11$ & $50.8 \pm 4.47$ & $55.7 \pm 1.97$  \\
        CS-Flow~\cite{Rudolph-WACV-2022}  & $90.9 \pm 0.83$ &  $87.3 \pm 0.58$ & $83.2 \pm 0.46$ & $98.5 \pm 0.28$ & $59.4 \pm 0.54$ & $68.4 \pm 0.42$ \\ 
        P-VQ~\cite{Kim-PR-2024}  &  {\color{orange}$\underline{94.3 \pm 0.23}$} &  $89.0 \pm 0.48$ & $79.2 \pm 0.04$ & - & - & - \\ 
        QFAE~\cite{dalmonte-wacv-2026}  &  {\color{orange}$\underline{94.3 \pm 0.18}$} &    {\color{orange}$\underline{91.8 \pm 0.55}$} &   {\color{orange}$\underline{83.8 \pm 0.46}$} & - & $65.5 \pm 1.96$ & -  \\
        MTL-MAD (Ours) & {\color{blue}$\mathbf{95.5 \pm 0.19}$} &  {\color{blue}$\mathbf{92.8 \pm 1.28}$} &  {\color{blue}$\mathbf{86.0 \pm 0.01}$} &  {\color{blue}$\mathbf{98.9 \pm 0.22}$} &  {\color{blue}$\mathbf{80.8 \pm 2.34}$} &  {\color{blue}$\mathbf{78.1 \pm 1.51}$} \\
        \bottomrule
    \end{tabular}
  }
    \label{tab:main_results}
    \vspace{-0.3cm}
\end{table}

\begin{table}[t!]
    \centering
    \caption{Ablation study on BraTS2021, highlighting the impact of each task on the final performance. The performance is measured in terms of image-level AUROC.}
    \label{tab:ablation}
    \fontsize{8}{9}\selectfont{
    \begin{tabular}{cccccc}
    \toprule
          $\;$MIM$\;$	& $\;\;\;\;\;\;$Jigsaw$\;\;\;\;\;\;$	& $\;\;$\textsc{DeMixUp}$\;\;$ &	Augmented AD & Generated AD  & AUROC \\
             \midrule
          \textcolor{ForestGreen}{\checkmark} & \textcolor{Red}{\xmark} & \textcolor{Red}{\xmark} & \textcolor{Red}{\xmark} & \textcolor{Red}{\xmark} & $65.54$ \\
           \textcolor{Red}{\xmark} & \textcolor{ForestGreen}{\checkmark}  & \textcolor{Red}{\xmark} & \textcolor{Red}{\xmark} & \textcolor{Red}{\xmark} & $84.38$  \\
           \textcolor{Red}{\xmark}  & \textcolor{Red}{\xmark} & \textcolor{ForestGreen}{\checkmark} & \textcolor{Red}{\xmark} & \textcolor{Red}{\xmark} & $87.77$  \\
           \textcolor{Red}{\xmark}  & \textcolor{Red}{\xmark} &  \textcolor{Red}{\xmark} & \textcolor{ForestGreen}{\checkmark} & \textcolor{Red}{\xmark} & $88.36$ \\
           \textcolor{Red}{\xmark}  & \textcolor{Red}{\xmark} &  \textcolor{Red}{\xmark}  &\textcolor{Red}{\xmark} & \textcolor{ForestGreen}{\checkmark} &  $90.72$ \\
            \midrule
           \textcolor{ForestGreen}{\checkmark}& \textcolor{ForestGreen}{\checkmark}& \textcolor{Red}{\xmark} &  \textcolor{Red}{\xmark}  &\textcolor{Red}{\xmark}  & $84.78$  \\
           \textcolor{ForestGreen}{\checkmark}& \textcolor{ForestGreen}{\checkmark}& \textcolor{ForestGreen}{\checkmark} &  \textcolor{Red}{\xmark}  &\textcolor{Red}{\xmark} &  $93.40$ \\
           \textcolor{ForestGreen}{\checkmark}& \textcolor{ForestGreen}{\checkmark}& \textcolor{ForestGreen}{\checkmark} &  \textcolor{ForestGreen}{\checkmark}  &\textcolor{Red}{\xmark}  & $93.87$  \\
           \textcolor{ForestGreen}{\checkmark}& \textcolor{ForestGreen}{\checkmark}& \textcolor{ForestGreen}{\checkmark} &  \textcolor{ForestGreen}{\checkmark}  &\textcolor{ForestGreen}{\checkmark} & $95.49$ \\
    \bottomrule
    \end{tabular}
    }
\vspace{-0.1cm}
\end{table}

\vspace{-0.1cm}
\noindent
\textbf{Results.} In Table~\ref{tab:main_results}, we present our main results, comparing MTL-MAD and several leading models from the literature using the complete BMAD benchmark. Our analysis reveals that MTL-MAD outperforms all existing methods on all six datasets, with no exception. MTL-MAD surpasses its competitors on both challenging datasets, such as BTCV+LiTs and CAMELYON16, as well as on nearly saturated datasets, namely OCT2017. We observe a significant improvement on the BTCV+LiTs dataset, where our model achieves absolute gains of over $8\%$ in terms AUROC over the strongest competitor. This result is particularly important, since BTCV+LiTs is the most challenging dataset in BMAD. Due to its difficulty, some recent methods struggle~\cite{dalmonte-wacv-2026} or do not even attempt~\cite{Kim-PR-2024} to perform anomaly detection on BTCV+LiTs. 

\begin{figure*}[!t]
\begin{subfigure}[t]{.49\textwidth}
  \centering
  \includegraphics[width=.85\linewidth]{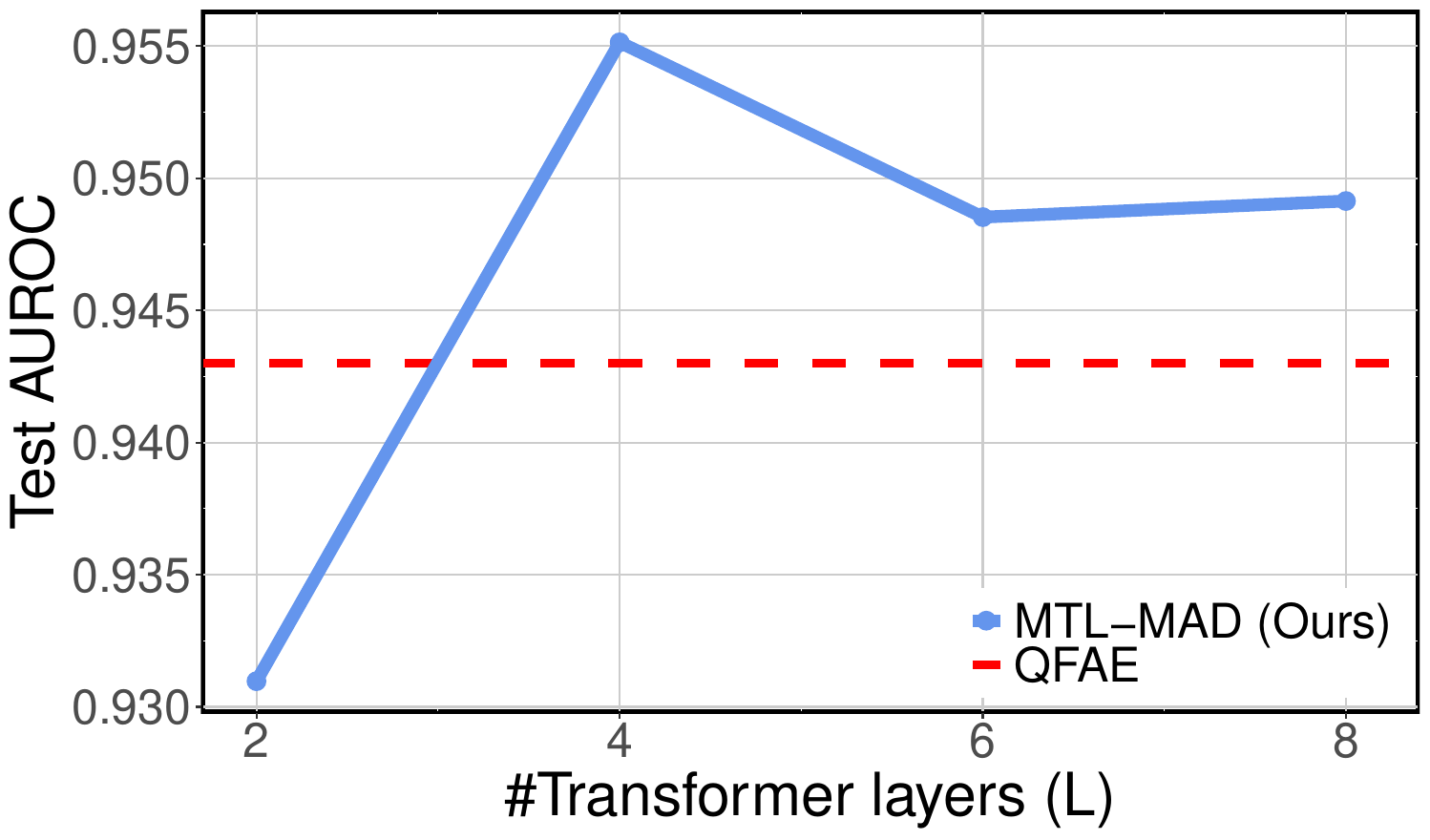}  
  \vspace{-0.2cm}
  \caption{Ablation for the depth $L$ of the shared encoder.}
  \label{fig:sub-L}
\end{subfigure}
\begin{subfigure}[t]{.49\textwidth}
  \centering
\includegraphics[width=.85\textwidth]{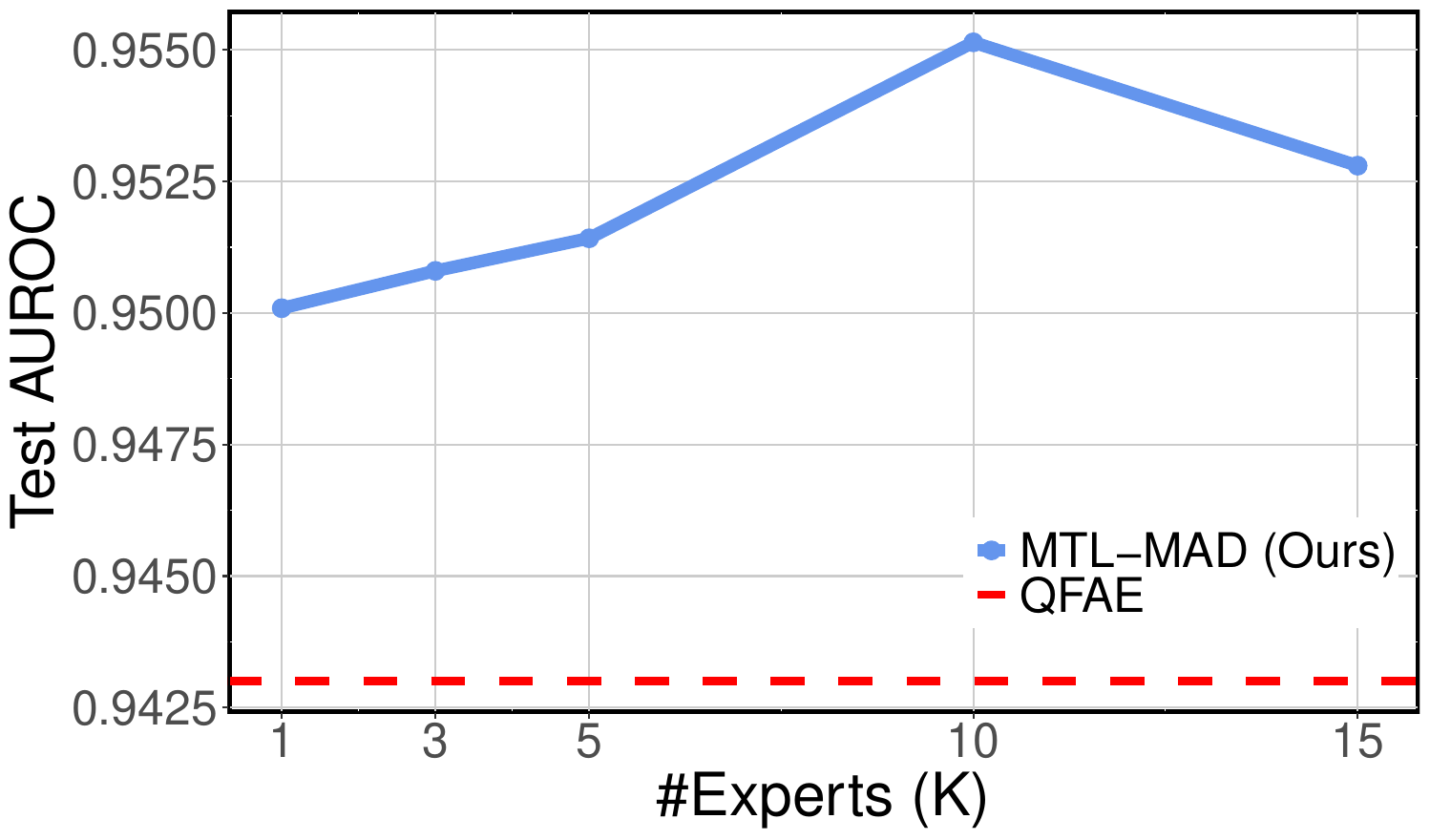}  
\vspace{-0.2cm}
  \caption{Ablation for the number of experts $K$ inside MoE.}
  \label{fig:sub-K}
\end{subfigure}
\vspace{-0.15cm}
\caption{Ablation study on BraTS2021 that showcases the importance of two hyperparameters, namely the number of transformer layers ($L$) and the number of experts ($K$). The performance is measured in terms of image-level AUROC. Best viewed in color.}
\vspace{-0.4cm}
\label{fig_ablation}
\end{figure*}

\vspace{-0.1cm}
\noindent
\textbf{Ablation studies.} In Table~\ref{tab:ablation}, we present an ablation study on BraTS2021, which highlights the contribution of each proxy task to the final performance. The results demonstrate the complementarity between the proposed proxy tasks. Moreover, our novel \textsc{DeMixUp} task yields one of the strongest individual performance levels, and contributes with the largest enhancement to the overall pipeline. The impact is clearly reflected in the evaluation measure, where the AUROC increases from $84.78\%$ to $93.40\%$ upon including \textsc{DeMixUp}.

In Figure~\ref{fig_ablation}, we present the impact of two important hyperparameters, namely the number of transformer layers (L) and the number of experts (K) within each layer, respectively. These ablation studies are conducted on BraTS2021. In both cases,  most of the considered values yield better performance than the previous state-of-the-art method~\cite{dalmonte-wacv-2026}. Moreover, in Figure~\ref{fig:sub-K}, we study scenarios where $K < 5$, and in all these cases, the experts are shared among tasks. The results indicate that there is some benefit of having distinct specialized experts per task. Since MTL-MAD comprises five tasks, we always utilize $K \geq 5$ to reach optimal performance in all our experiments.

\begin{figure}[t!]
\centering
  \includegraphics[width=1.0\textwidth]{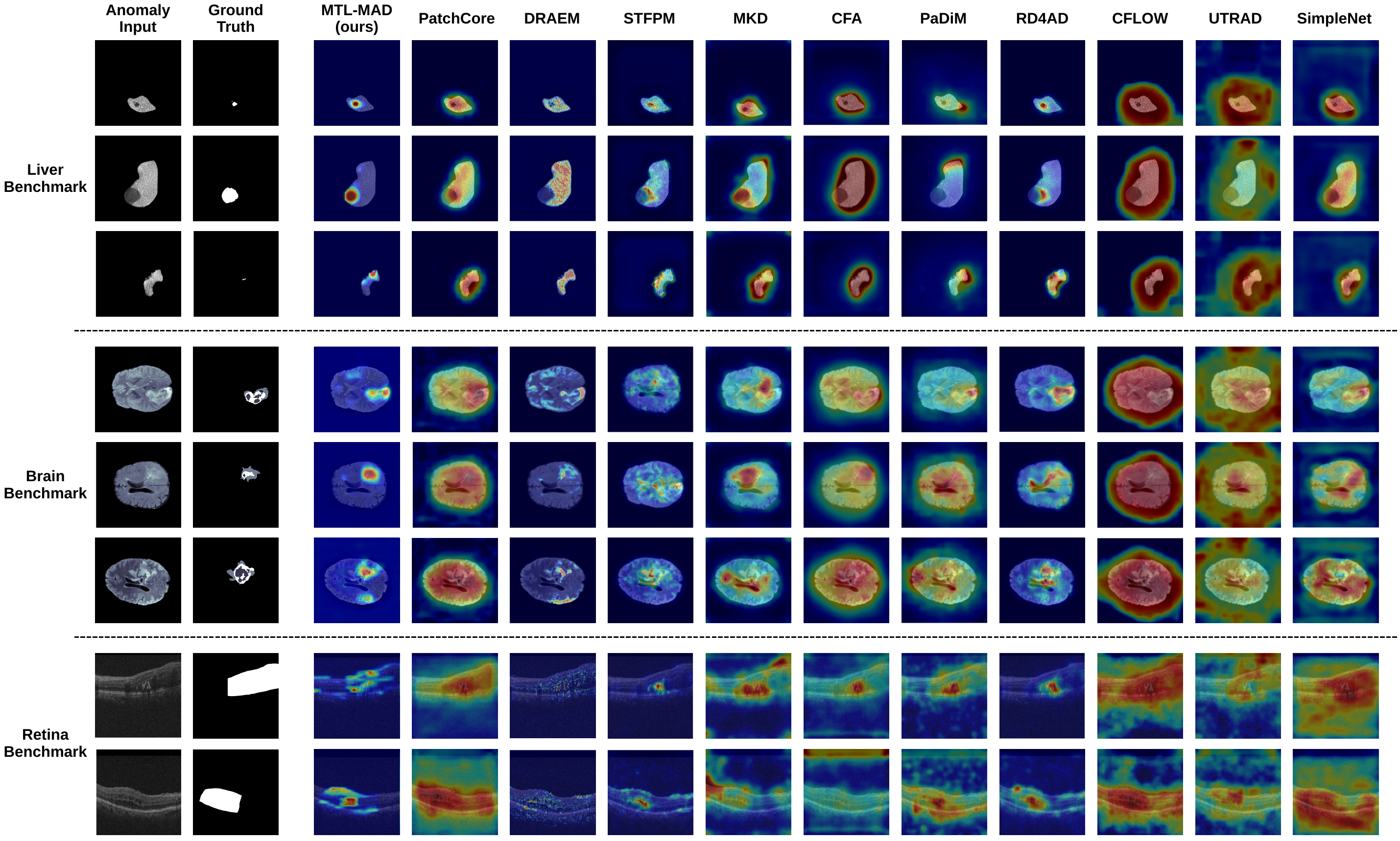}
  \vspace{-0.6cm}
  \caption{Qualitative visualizations of anomaly maps for the three benchmarks that have ground-truth segmentation maps: BTCV+LiTs, BraTS2021 and RESC. The examples are exactly those selected by Bao et al.~\cite{Bao-CVPRW-2024} for the BMAD introductory paper. Best viewed in color.}
  \label{fig:qual_res}
  \vspace{-0.3cm}
\end{figure} 

\vspace{-0.1cm}
\noindent
\textbf{Qualitative results.} 
In Figure \ref{fig:qual_res}, we present our anomaly maps for the examples illustrated in the BMAD paper~\cite{Bao-CVPRW-2024}, which allows us to compare our visualizations with many other methods. Our visual interpretation is carried out by an experienced physician, with a background in oncology and radiotherapy. 
Across all three evaluated benchmarks (BTCV+LiTs, BraTS2021 and RESC), the anomaly maps generated by MTL-MAD show a substantially more accurate visual correspondence with the pathological regions indicated by the ground-truth masks. Unlike the other models, MTL-MAD accurately delineates the tumor region, while significantly reducing false activations within healthy tissue. On liver CT, the proposed method identifies lesions with compact and well-defined localization, without the pathological over-extension observed in methods such as CFA, CFLOW, or UTRAD. On brain MRI, MTL-MAD preserves the normal anatomical structure, while selectively highlighting only the affected regions, demonstrating high sensitivity and specificity. In contrast, the other approaches tend either to underestimate lesion size or to generate diffuse anomaly maps that are difficult to interpret clinically. Similarly, on the retina benchmark, MTL-MAD provides the clearest separation between physiological and pathological areas, maintaining focal localization of anomalies, while minimizing background noise. These findings indicate that MTL-MAD maintains stable performance across different types of medical imaging modalities, including liver CT, brain MRI, and retinal OCT, demonstrating an increased capacity to adapt to multiple clinical and pathological contexts. Moreover, our method exhibits improved accuracy in detecting small lesions that are difficult to identify, as illustrated in both liver CT and brain MRI domains. The maps generated by MTL-MAD more closely reflect the true distribution of disease, which facilitates medical interpretation and future clinical integration. Another important advantage is the significant reduction in over-detection observed in competing methods, where entire anatomical structures are frequently marked as abnormal. By contrast, MTL-MAD maintains an optimal balance between sensitivity and localization precision. This performance demonstrates that the proposed method can extract clinically relevant pathological features without compromising the integrity of normal anatomical regions. Overall, MTL-MAD is not only capable of detecting the presence of disease, but also of providing a more faithful and clinically interpretable localization compared with the other methods.

\section{Conclusion}
\vspace{-0.1cm}

In this study, we introduced a multi-task learner for medical anomaly detection. By meticulously integrating complementary proxy tasks, MTL-MAD outperformed state-of-the-art methods on six datasets from the recently introduced BMAD benchmark. More importantly, we demonstrated exceptional performance levels without any model pre-training, further confirming that the power of MTL-MAD stems solely from the pretext tasks. By routing task-related tokens to specific experts of the MoE-based transformer, we obtained a versatile framework, able to bypass potentially conflicting loss terms via effective task decoupling embedded in the proposed architectural design. Finally, a subjective interpretation from a trained physician clearly indicated the precise anomaly localization abilities of MTL-MAD, which streamlines the subsequent manual diagnosis provided by physicians.

\vspace{-0.1cm}
In future work, we aim to integrate our anomaly detection framework into downstream diagnosis, to assist AI-based diagnosis tools in taking better informed decisions.

\section*{Acknowledgment}

This work was supported by a grant of the Ministry of Research, Innovation and Digitization, CCCDI - UEFISCDI, project number PN-IV-P7-7.1-PED-2024-1856, within PNCDI IV.

\bibliographystyle{plainnat}
\bibliography{refs}

\appendix

\section{Limitations}
\label{sec_limit}

While the proposed multi-task self-supervised framework demonstrates strong performance in medical anomaly detection, several limitations should be noted. First, the approach relies on proxy self-supervised and pseudo-labeling tasks. Taken alone, each task is only indirectly related to pathological abnormalities. As a result, there is an inherent risk of encountering mismatched anomalies and proxy tasks, where learned representations may not fully capture clinically relevant deviations from normal anatomy. This can potentially lead to reduced sensitivity for certain subtle or rare anomaly types, as well as false responses to irrelevant image variations. We mitigate such potential mismatches by aggregating multiple proxy tasks, so that mismatches caused by certain proxy tasks can still be recovered by other (well-aligned) proxy tasks. Hence, our design is based on a reasonable assumption, namely that proxy tasks are less likely to fail together, at exactly the same time.

Second, multi-task learning introduces the possibility of task interference. Competing objectives may lead to gradient conflicts or uneven optimization across tasks, where some tasks dominate the shared representation, while others contribute marginally. This can result in suboptimal feature learning compared to more carefully isolated single-task models. Our attempt to mitigate this limitation is to employ Mixture-of-Experts (MoE) and decouple the tasks by routing task-specific tokens to different experts. 

Third, the proposed framework increases computational complexity during training. The inclusion of multiple tasks naturally leads to increased training cost. We try to counter this effect by using a relatively lightweight shared encoder, with at most $L=12$ transformer blocks. As reference, both ViT-Small and ViT-Base contain $L=12$ transformer blocks \cite{Dosovitskiy-ICLR-2021}.


Finally, the performance of MTL-MAD may vary depending on the number of experts and depth of the shared encoder. To obtain optimal results in other medical imaging domains, hyperparameters should be established on a small BMAD-style validation set that can easily be collected in practice.


\begin{table}[t!]
    \centering
    \caption{Optimal hyperparameter configurations for MTL-MAD, determined via grid search on the official validation splits.}
    \label{tab:hyperparams}
    \begin{tabular}{lcccc}
    \toprule
        Dataset & Image resolution & $L$ & $d$ & $K$\\
             \midrule
        BraTS2021 & $256 \times 256$ & $4$ & $128$ & $10$ \\
        RESC & $256 \times 256$ & $4$ & $128$  & $5$ \\
        RSNA & $256 \times 256$ & $4$ & $128$  & $5$ \\
        OCT2017 & $256 \times 256$ & $4$ & $128$  & $5$ \\
        BTCV+LiTs & $256 \times 256$ & $12$ & $768$ & $10$ \\
        CAMELYON16 & $256 \times 256$ & $12$ & $768$ & $10$ \\
    \bottomrule
    \end{tabular}
\vspace{-0.1cm}
\end{table}

\section{Broader Impact}
\label{sec_broad_impact}

This work has a high potential of generating a positive impact in healthcare by improving anomaly detection in medical images, which may support early diagnosis and reduce clinical workload. However, deployment in clinical settings may also introduce several risks. Model errors (false positives or false negatives) could lead to missed diagnoses or unnecessary interventions, particularly if clinicians over-rely on automated outputs. Additionally, performance may vary across demographic groups or imaging settings, raising fairness concerns if training data is not representative. Privacy is another key consideration, as medical imaging data is highly sensitive. To mitigate these risks, systems based on our work should pass thorough clinical validation, to ensure diverse training data and secure data handling practices. 

Since the use of multi-task learning and Mixture-of-Experts increases computational cost during training, our approach may contribute to a higher energy consumption and a negative environmental impact. For limited training resources, we thus recommend checking if methods based on a single proxy task provide good-enough results, before considering multi-task learning.

\begin{table}[t!]
    \centering
    \caption{List of prompts employed for synthetic anomaly generation. For each medical domain, we randomly sample one prompt from the list to generate diverse pseudo-anomalies.}
    \label{tab:prompts}
    \setlength\tabcolsep{0.12cm}
    \fontsize{8}{9}\selectfont{
    \begin{tabular}{lp{11.5cm}}
    \toprule
        Medical  & \multirow{2}{*}{Prompts}\\
        domain & \\
        \midrule
        \multirow{5}{*}{Brain MRI} & > Brain lesion\\ 
        & > White area, high intensity white blob consistent with the surrounding area, white matter lesion \\ 
        & > Hyperintense spot \\ 
        & > Brain tumor \\ 
        & > Bright white spot\\
        
        \midrule
        \multirow{12}{*}{Liver CT} & > Generate a solid, hypervascular nodule in the masked region.\\ 
        & > Show bright arterial phase hyperenhancement and portal venous phase washout. \\
     & > Inpaint a hypodense lesion that shows peripheral, nodular enhancement in the arterial phase and gradually fills in centripetally (from outside-in) on the portal venous phase.\\
    & > Generate a hypodense, ring-enhancing lesion in the masked area, most visible in the portal venous phase.\\
    & > A simple hepatic cyst. The lesion must be sharply defined, non-enhancing, and hypodense (near-water density, 0-20 HU).\\
    & > Inpaint the masked liver parenchyma to be hypodense (darker) than the spleen, consistent with hepatic steatosis.\\
    & > Add an anomaly to the liver image. \\
    & > Inpaint black blobs. \\
        
        \midrule
        \multirow{6}{*}{Retinal OCT} & > Add a tightly packed cluster of bright, waxy dots with very sharp, distinct edges.\\
        & > Draw several irregular, dark crimson splatters and deep dots that look like wet ink stains on the surface.\\
        & > Create a few soft, fluffy, pale white patches with blurred and faded edges, resembling thin clouds.\\
        & > Generate a chaotic, tangled web of extremely thin, bright, thread-like lines that branch out irregularly.\\
        & > Create a large, sharply outlined area where the  background looks faded, revealing pale and \\
        &exposed thick white vessels underneath.\\
        
        \midrule
        \multirow{8}{*}{Chest X-ray} &  > Create multiple small, calcified nodules consistent with old granulomatous disease. \\
     & > Airspace consolidation with air bronchograms, consistent with pneumonia, in the masked region. \\
    & > Opacities, suggestive of atypical pneumonia.\\
    & > Pneumothorax in the masked apex, showing a visible visceral pleural line and an absence of lung markings.\\
    & > Create a pleural effusion in the masked area, characterized by blunting of the costophrenic angle and a meniscus sign.\\
    & > Add an anomaly to the chest image. \\
    
    \midrule
        \multirow{8}{*}{Histopathology} &  > Cells exhibit nuclei of vastly different sizes and irregular, angular shapes. \\
        & Some nuclei are hyperchromatic (deep purple) with thickened nuclear membranes.\\
        & > Large, distinct, reddish-purple dots are visible inside the enlarged, pale nuclei.\\
        & > Histological image showing multiple atypical mitotic figures.\\
        & Visible tripolar and quadripolar spindles (star-shaped or multi-pronged dark chromatin clumps) indicating rapid, disorganized cell division.\\
        & > Cells are haphazardly arranged and `piling up' rather than forming organized layers. \\
        & > Cells are highly primitive, bizarre, and unrecognizable compared to normal tissue.\\
    \bottomrule
    \end{tabular}
    }
\vspace{-0.1cm}
\end{table}

\section{Hyperparameter Configuration and Implementation Details}
\label{sec_more_hyper}

In Table~\ref{tab:hyperparams}, we report the optimal hyperparameter configuration for each of the six datasets. We employ larger backbones ($L=12$, $d=768$) for the more complex BTCV+LiTs and CAMELYON16 datasets, while keeping the backbone fairly lightweight ($L=4$, $d=128$) for the other datasets.

\begin{figure}[t]
    \centering
    \includegraphics[width=1\linewidth]{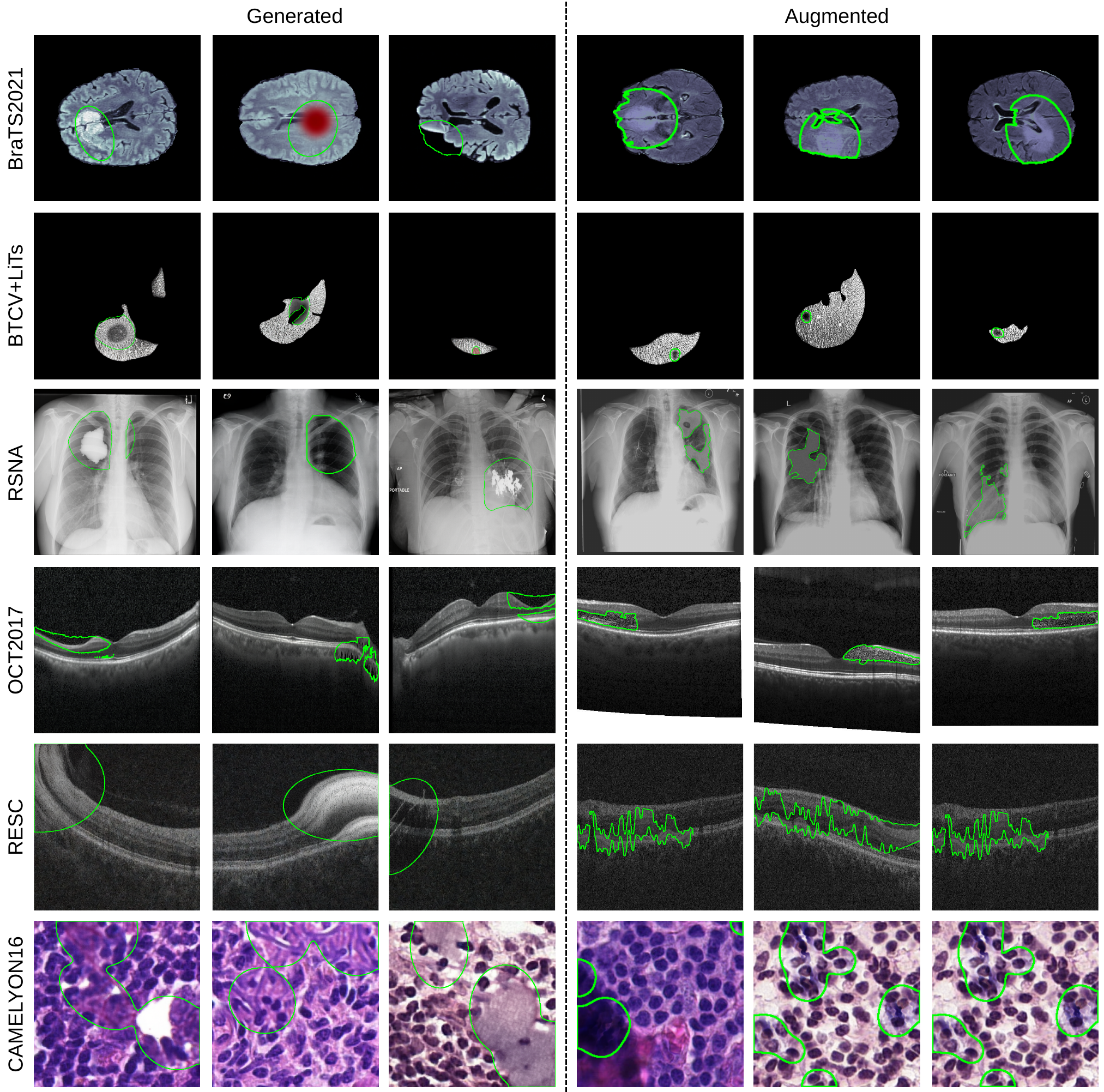}
    \caption{Examples of pseudo-anomalies generated via VLMs (left-hand side) or data augmentation (right-hand side), which are used by our pipeline in the pseudo-anomaly classification tasks (Task 4 and Task 5). The full sets of generated and augmented anomalies are released along with the code. Best viewed in color.}
    \label{fig:anomalies}
\end{figure}

In Table~\ref{tab:prompts}, we show the text prompts used to generate synthetic anomalies via frozen VLMs. To generate a pseudo-anomaly for a certain medical domain, we uniformly sample one prompt from the corresponding list of prompts. In Figure \ref{fig:anomalies}, we showcase several examples of pseudo-anomalies obtained via generative VLMs or data augmentations, respectively. While pseudo-anomalies are sometimes too mild (hard to spot) or unrealistic (e.g.~colorful spots in grayscale images), the corresponding pseudo-anomaly classification tasks remain a critical addition to the success of MTL-MAD.

\begin{figure*}[!th]
\centering
  \includegraphics[width=.5\linewidth]{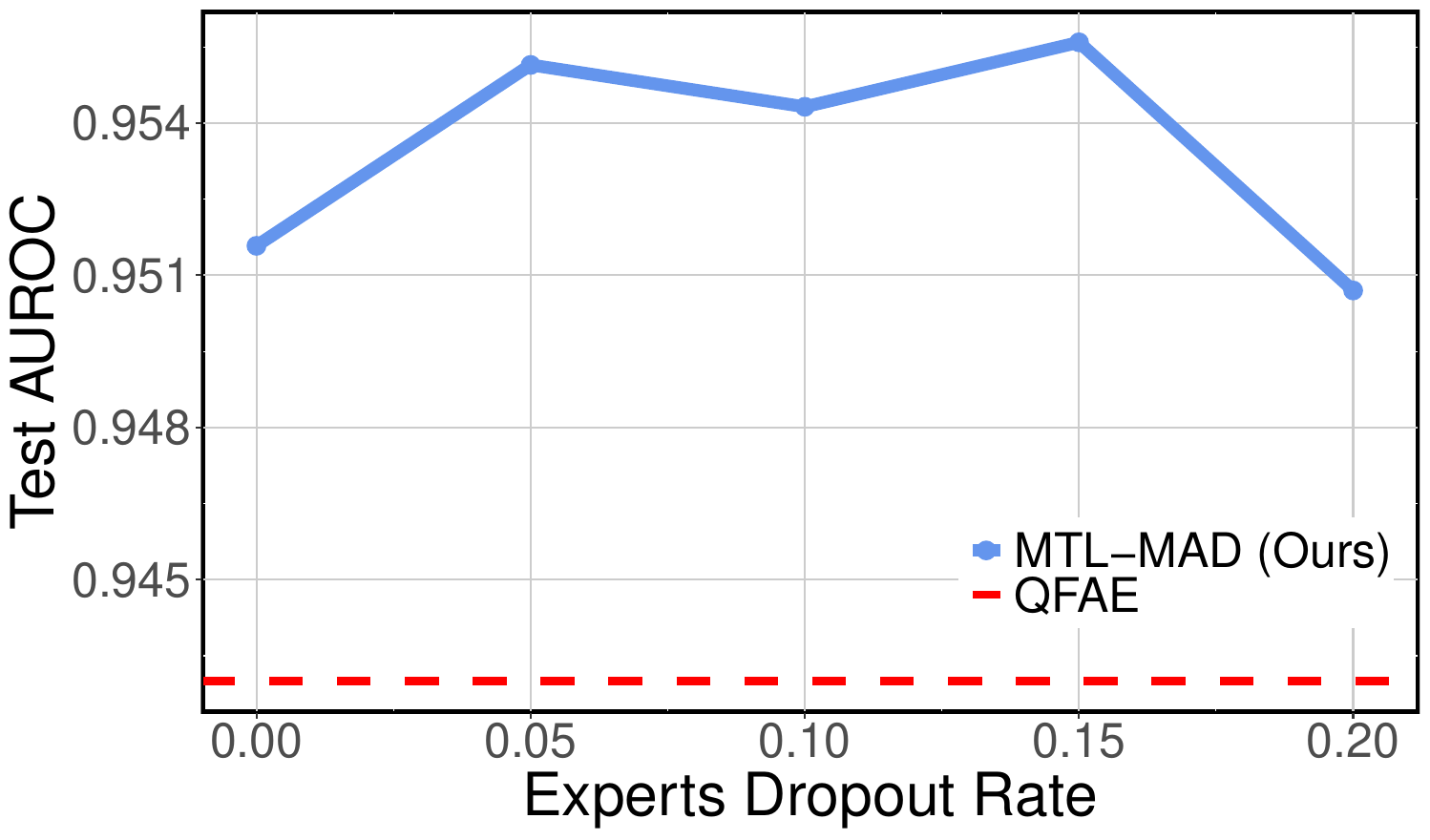}  
  \caption{Impact of varying expert dropout rates on performance. The study is conducted on BraTS2021.}
\label{fig_dropout_rate}
\end{figure*}

In Figure~\ref{fig_dropout_rate}, we present the impact of varying the expert dropout rate on the final performance. Notably, dropout rates between $5\%$ and $15\%$ yield a performance improvement over the no-dropout baseline.

\section{Compute Environment}
\label{sec_compute}

The experiments have been conducted on four local nodes with different configurations. The best-equipped node has a single NVIDIA A100 80GB PCIe GPU, 64 GB of memory and a CPU featuring 56 physical cores. The least capable configuration is equipped with a single NVIDIA RTX 3090 24 GB, 64 GB of system memory and 10 physical cores.



\end{document}